\documentclass{article}
\usepackage{arxiv}

\usepackage{graphicx} 
\usepackage{hyperref}
\usepackage{booktabs}
\usepackage{amsmath}
\usepackage{amsfonts}
\usepackage{pdflscape}
\usepackage{algorithm}
\usepackage{algorithmicx}
\usepackage[noend]{algpseudocode}
\usepackage{natbib}

\DeclareFixedFont{\ttb}{T1}{txtt}{bx}{n}{10} 
\DeclareFixedFont{\ttm}{T1}{txtt}{m}{n}{10}  

\usepackage{color}
\definecolor{deepblue}{rgb}{0,0,0.5}
\definecolor{deepred}{rgb}{0.6,0,0}
\definecolor{deepgreen}{rgb}{0,0.5,0}

\usepackage{listings}

\lstdefinestyle{python}{
    language=Python,
    basicstyle=\ttm,
    otherkeywords={self},             
    numbers=left,
    stepnumber=1,
    keywordstyle=\bfseries\color{deepblue},
    commentstyle=\color{deepgreen}\itshape,
    stringstyle=\color{deepgreen},
    emph={Budget,build_portfolio,Solver,run_sequence},          
    emphstyle=\color{deepred},    
    stringstyle=\color{deepgreen},
    frame=tb,                         
    showstringspaces=false,           %
    numbersep=3pt,
    captionpos=b,
    escapeinside={<!}{!>}
}

\title{Exploiting Configurations of MaxSAT solvers}
\author{
    Josep Alòs \\
    Logic \& Optimization Group (LOG), University of Lleida, Spain \\
    \texttt{josep.alos@udl.cat}
    \And
    Carlos Ansótegui \\
    Logic \& Optimization Group (LOG), University of Lleida, Spain \\
    \texttt{carlos.ansotegui@udl.cat}
    \And
    Josep M. Salvia \\
    Logic \& Optimization Group (LOG), University of Lleida, Spain \\
    \texttt{jsh7@alumnes.udl.cat}
    \And
    Eduard Torres \\
    Logic \& Optimization Group (LOG), University of Lleida, Spain \\
    \texttt{eduard.torres@udl.cat}
}
\date{June 2023}

\renewcommand{\headeright}{}
\renewcommand{\undertitle}{}
\renewcommand{\shorttitle}{Exploiting Configurations of MaxSAT solvers}

\hypersetup{
pdftitle={Exploiting Configurations of MaxSAT solvers},
pdfsubject={cs},
pdfauthor={Josep Alòs, Carlos Ansótegui, Josep M. Salvia, Eduard Torres},
pdfkeywords={maximum satisfiability, maxsat evaluation, automatic configuration},
}

\begin{document}

\maketitle

\begin{abstract}

In this paper, we describe how we can effectively exploit alternative parameter configurations to a MaxSAT solver. We describe how these configurations can be computed in the context of MaxSAT. In particular, we experimentally show how to easily combine configurations of a non-competitive solver to obtain a better solving approach.

\end{abstract}

\section{Introduction}\label{sec:introduction}

Since 2006, the MaxSAT Evaluation (MSE)~\cite{bacchus_maxsat_2022} has been held annually with the primary objective of advancing MaxSAT technology and assessing its current state-of-the-art. The evaluation consists of multiple solvers being tested on various benchmarks across different evaluation tracks. This event has undeniably spurred the MaxSAT community to create more cutting-edge solvers and enhance their competitiveness.

It is not surprising that solver performance depends on several factors, including the \emph{power} of the algorithm implemented by the solver, proper configuration of solver parameters to unleash its full potential, and implementation issues. Therefore, we must interpret the MaxSAT Evaluation ranking results carefully and derive conclusions according to the goal of our analysis. For example, a similar or \emph{weaker} algorithm could outperform other approaches thanks to better implementation of data structures or a previous tuning process of its input parameters.

From an industrial point of view, we mainly care about obtaining an effective solving approach that is ready for deployment for a particular problem subject to available resources (computing power, environment restrictions, licenses available, etc). From a research point of view, we are more interested in identifying the potential of new solving approaches that lead to further promising research avenues.

Our aim is to satisfy both industrial and research perspectives by identifying the best possible solving approach that can be achieved from a single solver while adhering to certain restrictions. In particular, we treat the solver as a black box, meaning that we cannot access its source code, nor do we have any domain knowledge of the problem to be solved, meaning that we cannot utilize any specific structure feature.

Despite these constraints, our approach enables us to unleash the hidden potential of the solver and avoid incorrect rankings of better algorithms that have not been appropriately configured or restarted. Additionally, our study emphasizes the importance of being cautious when interpreting rankings based on the MaxSAT Evaluation, as we mentioned previously.

In this paper, we first show how to effectively configure MaxSAT solvers using Automatic Configuration (AC) tools (tuners), specifically GGA~\cite{AnsoteguiST09} and SMAC~\cite{lindauer_smac3_2021}. Then, we show that we can take advantage of not only the best configuration returned by the tuners but also a selection of the configurations seen by the tuner during the AC process. With these configurations, we can then build a simple portfolio that runs in parallel these configurations if enough computational resources are available.

We also demonstrate how to create a sequential portfolio that schedules the execution of different parametrizations of a single MaxSAT solver on a given number of cores within a specified timeout. This approach can be thought of as a restarting strategy, where a different configuration of the solver parameters is selected at each restart. 

It is worth mentioning that all these approaches are agnostic of the structure of the instances. Otherwise, we should explore extending other approaches available in the literature such as ISAC++~\cite{kadioglu_isac_2010}.

Finally, we integrate all these building processes in the OptiLog framework~\cite{alos_optilog_2022}. With the new APIs, the user can provide an input MaxSAT solver and its parameters through the \emph{BlackBox Module}, and OptiLog automatically generates a new solving approach for a given number of cores.

We conducted an extensive experimental investigation on the Weighted Incomplete track of the MaxSAT Evaluation 2022, with a particular focus on the highly configurable MaxSAT solver Loandra~\cite{berg_loandra_2019}. In this track, Loandra ranked sixth when restricted to a timeout of 60 seconds. Our approach involves the construction of parallel and sequential portfolios based solely on Loandra, which significantly improves its performance.

\section{Preliminaries}\label{sec:preliminaries}

MaxSAT is the optimization variant of the SAT decision problem. While for SAT the goal is to find an assignment to the Boolean variables (solution) that satisfies all the clauses in the input CNF formula, in MaxSAT we look for a solution that satisfies the maximum possible number of clauses. Since some of these clauses can be falsified we refer to them as \emph{soft} clauses. Within the MaxSAT community, it is typical to reformulate the problem from a minimization perspective aiming to find a solution that falsifies the minimum possible number of soft clauses. 

There are several variants of the MaxSAT problem. We can add weights to the soft clauses that represent the cost of falsifying the clause. In this case, we want to look for a solution that minimizes the aggregated cost of the \emph{Weighted} soft clauses. Additionally, we can have \emph{hard} clauses, i.e., clauses that cannot be falsified by the solution.

MaxSAT solvers have experimented a great success in the last decade. Among these solvers, we find complete (or exact) solvers and incomplete solvers. Complete solvers provide optimal solutions while incomplete solvers report solutions as good as possible, but are not required to guarantee their optimality. These solvers can either refine a lower bound (lb) on the cost of the optimal solutions or an upper bound (ub), or both. In particular, incomplete solvers iteratively report (whenever possible) a better (smaller) upper bound on the optimal solution. 

\section{The MaxSAT Evaluation}\label{sec:mse}

The MaxSAT Evaluation 2022 was structured into three tracks: main track complete (unweighted and weighted variants), main track incomplete (unweighted and weighted variants), and the special incremental MaxSAT track. In this paper, we focus on the incomplete track for weighted MaxSAT instances with a timeout of 60 seconds.

The term \emph{incomplete} refers to the type of MaxSAT solvers which are not required to be exact, i.e., they do not need to certify the optimum. Their goal is to report the best possible solution within a given timeout.
The term \emph{weighted} refers to the variant of MaxSAT instances. The weighted MaxSAT variant allows integer weights for the soft clauses plus the hard clauses.

We consider the timeout of 60 seconds useful for our study since it is a realistic scenario of industrial applications where we require a suboptimal solution in a short time window and because our automatic configuration process, given the computational resources we have available, can be restricted to two days (see Section~\ref{sec:configuring}).

The MaxSAT Evaluation 2022 incomplete (weighted) track involved 197 MaxSAT instances and 10 incomplete solvers: DT-HyWalk~\cite{zheng2022decision}, noSAT-MaxSAT~\cite{lubke2022nosat}, NuWLS-c~\cite{chu2022nuwls}, Exact~\cite{elffers_divide_2018, jo_devriendt_exact_2023}, Loandra~\cite{berg_loandra_2019}, Open-WBO-inc (two variants)~\cite{DBLP:journals/jsat/JoshiKRM19}, and TT-Open-WBO-Inc (three variants)~\cite{nadel_polarity_2020}.

Each solver $s$ was ranked according to the scoring function $score(s)$ shown in Equation~\ref{eq:score_mse:score}.

\begin{equation}\label{eq:score_mse:score}
score(s) = \frac{\sum_{i=1}^{i=n} score(s,i)}{n}
\end{equation}

Given a set of $n$ instances, the $score(s)$ of a MaxSAT solver $s$ is the average of the scores for each instance, computed by $score(s, i)$ in Equation~\ref{eq:score_mse:score_inst}.

\begin{equation}\label{eq:score_mse:score_inst}
score(s,i) =  \frac{1+ \mbox{best-known ub for instance } i}{1+ \mbox{ub for }i \mbox{ found by }s}
\end{equation}

The $score(s, i)$ function computes the ratio between the \emph{best-known upper bound} of an instance $i$ and the bound reported by the solver $s$ on the same instance. Assuming that $(best$-$known\ ub) \leq ub$ (which is the case for the MaxSAT Evaluation), the computed value ranges between 0 and 1, where higher values correspond to better upper bounds.

The competition has some specific rules about what is and is not allowed in the implementation of the solvers. In particular, the solvers are not allowed to employ triggers to modify their behavior, which is deemed to be specific to particular instances. However, solvers can concatenate the usage of different solving techniques.

In the most recent competition, the MaxSAT solvers used a variety of strategies and solvers. Some of these solvers are outlined below, along with the various approaches they employ in order to find improved solutions.

\begin{enumerate}
    
    \item \textbf{NuWLS-c}: This solver adopts two solvers, the NuWLS solver, which is an improvement of SATLike, and the integration of TT-Open-WBO-Inc.
    \item \textbf{TT-Open-WBO-Inc}: This solver uses four different strategies, including SATLike for inprocessing, a modified version of Mrs. Beaver for unweighted instances, BMO-clustering for weighted instances, and Polosat, a SAT-based local search method. This MaxSAT solver has three different distributions:
    \begin{itemize}
    \setlength{\itemindent}{.25in}
        \item \textbf{(g)} Which incorporates the Glucose 4.1 SAT Solver.
        \item \textbf{(i)} Which incorporates the new Intel SAT Solver.
        \item \textbf{(is)} Which incorporates the new Intel SAT Solver and is tuned for short invocations.
    \end{itemize}
    
    \item \textbf{DT-HyWalk}: This solver employs three distinct strategies, including a direct call to a SatSolver, the SATLike solver for local search, and the use of another MaxSAT solver, TT-Open-WBO-Inc.
    \item \textbf{Loandra}: This solver utilizes two core algorithms, namely a Core-Based algorithm and a Linear algorithm.
\end{enumerate}

Table~\ref{tab:mse}, column ``MSE'', shows the results of the MaxSAT Evaluation 2022 for the top six solvers at the incomplete weighted track with 60 seconds timeout. As we can see, the MaxSAT solver Loandra was not competitive within this category. In this paper, we propose an approach that is agnostic of the structure of instances and only allows the usage of alternative configurations of the same input MaxSAT solver. We experimentally show the goodness of our approach on the MaxSAT solver Loandra.

\subsection{Reproducing the MaxSAT Evaluation for the Incomplete track}\label{sec:mse:reproducing}

All of our executions of the MaxSAT solvers are run on a computation cluster composed of nodes with two AMD 7402 processors (each with 24 cores at 2.8 GHz) and 21 GB of RAM per core, managed by Sun Grid Engine (SGE). All the experiments are managed using the \emph{Running Module} of the OptiLog framework.

Each execution is given 60s of CPU time and 32 GB of memory. As the memory requirements exceed the memory per core available, two slots are reserved and an affinity mask is set by SGE to restrict the execution to only one of the two cores. In contrast to the MaxSAT evaluation, each solver was evaluated with 50 different random seeds and we report results on the average score, and in some of the experiments, we also show the minimum and maximum scores, and the standard deviation.  

In the course of developing our experiments, we detected two problems with some executions of the solvers: 1) some executions report a bound that does not correspond to the real cost of the solution reported, and 2) some executions report a solution that does not satisfy the \emph{hard} clauses. 

To address these issues we conduct a validation step executed after the solver exhausts the 60s of CPU time. In particular: 

For 1), we trust the cost we compute from the solution reported, ignoring the bound reported by the solver.

For 2), we consider the solver was not able to find any solution at all. 

This validation step is also conducted during the automatic configuration process when we evaluate a particular configuration of the solver on a given instance (see Section~\ref{sec:configuring}).

The score for each solver is computed using the MaxSAT evaluation rules. In particular, it is important to define which is the set of best-known upper bounds that we use to compute the score. Table~\ref{tab:mse} shows in column ``MSE 2022'', the scores reported in the MaxSAT Evaluation 2022.

The rest of the columns present the results of the experimentation we conducted (in our cluster) using different sets of \emph{best-known upper bounds}. ``$VBS_b$'' uses the upper bounds found by the Virtual Best Solver of the solvers we executed, ``$MSE_b$'' uses the set of best-known upper bounds provided by the MaxSAT Evaluation, and ``$LRUNS_b$'' (Long Runs) uses a set of new \emph{best-known upper bounds} we computed by running Loandra and NuWLS-c (both with the default parameters) with a timeout of 12 hours. We recall that the score presented is the average score on 50 seeds in contrast to the MSE results where only 1 seed is used. As we can observe in Table~\ref{tab:mse}, the relative ranking of the solvers is preserved although we can observe variations in the scores reported. We will use in the rest of the paper the best bounds from $MSE_b$ + $VBS_b$ + $LRUNS_b$.

\begin{table}[ht]
\centering
\caption{Results of the MaxSAT Evaluation 2022 on the incomplete weighted track (60 seconds timeout) and reproduction of the Evaluation in our system with different sets of Best-Known Upper Bounds.}
\label{tab:mse}
\begin{tabular}{l|r|ccc}
\toprule
                            & MSE 2022 & \multicolumn{3}{c}{MSE 2022 on our system} \\
\midrule
Best-known UBs          & $MSE_b$      & $VBS_b$ & $VBS_b + MSE_b$ & $VBS_b + MSE_b + LRUNS_b$ \\
\midrule
Solvers                     &          & & & \\
\midrule
NuWLS-c                     & 0.759    & 0.7831     & 0.7590           &  0.7524    \\
DT-Hywalk                   & 0.732    & 0.7625     & 0.7414           &  0.7351    \\
TT-Open-WBO-inc (g)         & 0.728    & 0.7412     & 0.7221           &  0.7164    \\
TT-Open-WBO-inc (is)        & 0.726    & 0.7354     & 0.7201           &  0.7141    \\
TT-Open-WBO-inc (i)         & 0.720    & 0.7354     & 0.7178           &  0.7118    \\
Loandra                     & 0.693    & 0.7107     & 0.7003           &  0.6953    \\
\bottomrule
\end{tabular}
\end{table}

\section{Automatic Configurators (AC)}\label{sec:ac}

In this section, we review the Automatic Configuration Problem and two state-of-the-art automatic configuration algorithms or tuners. 

\subsection{The Automatic Configuration Problem}\label{sec:ac:ac_problem}

Given a target algorithm $A$ with parameters $\{p_1, \ldots, p_n\}$
of domain $d(p_i)$. We define the parameter space $\Theta$ of $A$ as
the subset $d(p_1) \times \ldots \times d(p_n)$ of \emph{valid} parameter
combinations. Depending on the parameter, $d(p_i)$ can be categorical,
a discrete domain of fixed values with no predefined order, or numerical,
which represent integer or real values.
Then, we define the \emph{Automatic Algorithm Configuration} (AAC) problem 
as the optimization problem that consists of exploring $\Theta$ to find a configuration 
$\theta \in \Theta$ of $A$, which given a set of problem instances
$\Pi$, minimizes a cost metric $\hat{c} : \Theta \times \Pi \rightarrow \mathbb{R}$,
without exceeding a configuration budget $B$.

\begin{figure}[ht]
    \centering
    \includegraphics[width=.65\textwidth]{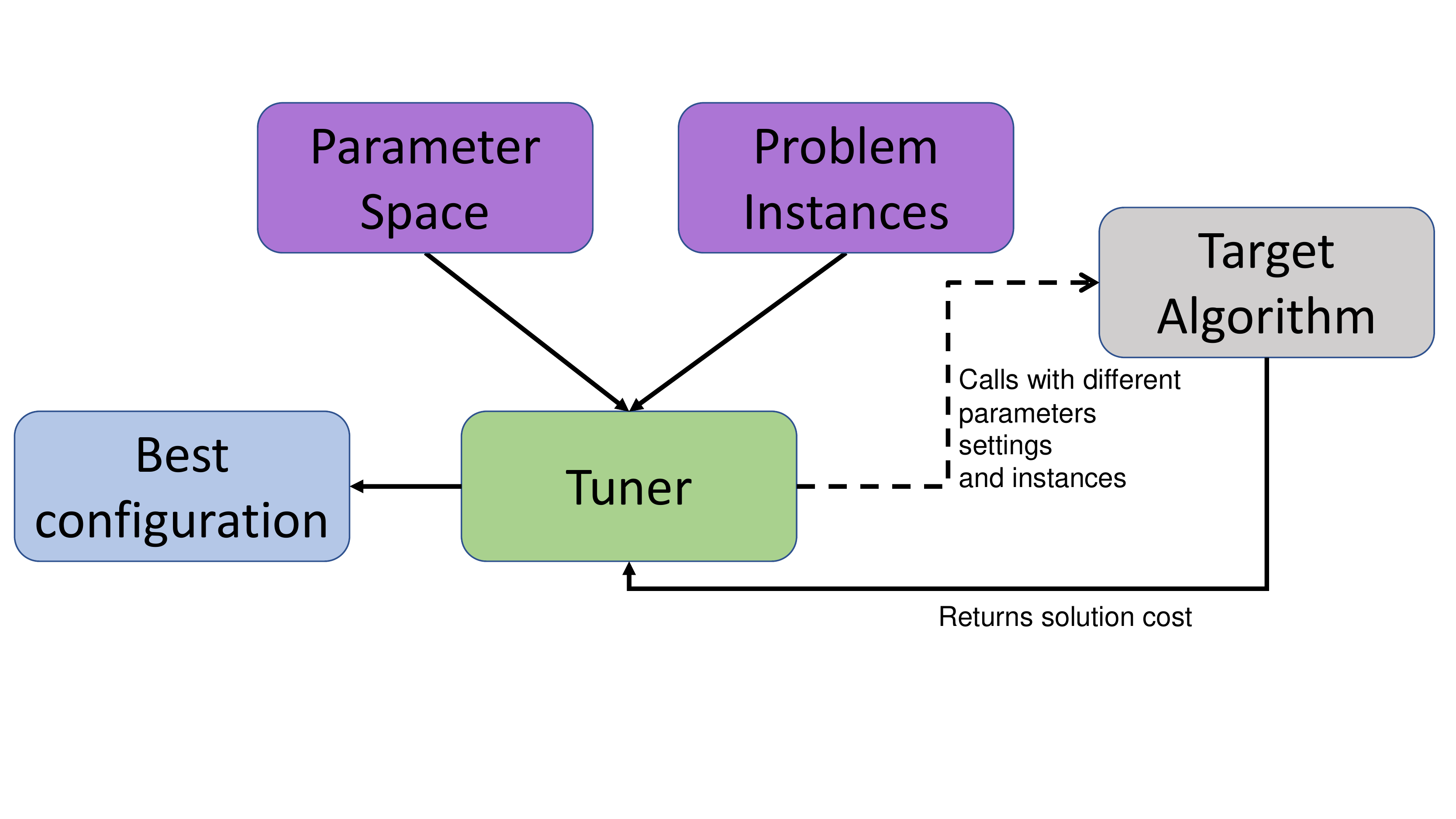}
    \caption{Visualization of the Automatic Configuration process.}
    \label{fig:aac-overview}
\end{figure}

It is common for $A$ to be a black box (target algorithm), meaning it accepts some inputs (the parameters) and provides some output (e.g., $\hat{c}$), but we cannot see its internal functionality. This allows AAC to generalize to any type of algorithm but makes it more challenging for algorithm tuners since they cannot use $A$ to infer additional information about $\Theta$. In practice, $A$ is implemented as a binary file that outputs its results in a format suitable for its domain but may not be suitable for the AAC tool. Moreover, it may be necessary to limit the resources that $A$ can use to solve an instance, such as memory or CPU time. The standard way of addressing these issues in AAC tools is for the user to replace $A$ with a wrapper script that handles these and any other necessary aspects. Figure~\ref{fig:aac-overview} describes the automatic configuration process where the tuner is a solver for the AAC problem.

\subsection{The GGA Automatic Configurator}\label{sec:ac:gga}

The Gender-Based Genetic Automatic Algorithm Configuration (GGA) is a genetic algorithm that was introduced in \cite{AnsoteguiST09} to search for high-quality configurations. It was one of the pioneering algorithms that supported continuous parameters and introduced the novel concept of \emph{gender} to apply diverse selection pressures to the population's individuals.

\begin{algorithm}[ht]
    \begin{algorithmic}[1]
        \small
        \renewcommand{\algorithmicrequire}{\textbf{Input:}}
        
        \Require {Target Algorithm $A$, Parameter Space $\Theta$, Instances $\Pi$,
                  Performance Metric $\hat{c}$,
                  \# MiniTournaments $N$,
                  Configuration Budget $B$
                  }
        \Function{GGA}{$A, \Theta, \Pi, \hat{c}, N, B$}
        \State pop $\gets$ initPopulation($\Theta$)\label{alg:gga:init-pop}
        \State $j = 0$
        \While {$B$ not exhausted \textbf{and} threshold not achieved}\label{alg:gga:main-loop}
            \State $j = j + 1$
            \State $\Pi_{j}$ $\gets$ selectInstances($\Pi$, $j$)\label{alg:gga:inst-selection}
            \State <$w_1, ..., w_{N}$> $\gets$ runMiniTournaments($A$, pop.comp, $\Pi_{j}$, $\hat{c}$, pop.comp/$N$)\label{alg:gga:mini-tournament}
            \State offspring $\gets$ applyCrossoverAndMutate(pop.noncomp, <$w_1, ..., w_{N}$>, $\Theta$)\label{alg:gga:offspring}
            \State pop $\gets$ agingAndDeath($w_{1}$, pop) $\cup$ offspring\label{alg:gga:aging}
        \EndWhile
        \Return $w_{1}$
        \EndFunction
    \end{algorithmic}
    
    \caption{GGA}
    \label{alg:gga}
\end{algorithm}

Algorithm~\ref{alg:gga} shows the pseudocode of the GGA algorithm, which takes as input the target algorithm $A$, its parameter space $\Theta$, a set of training instances $\Pi$, a performance metric $\hat{c}$ to optimize (e.g., time, accuracy, quality within a fixed timeout, etc), the number $N$ of GGA mini-tournaments (which will be explained shortly), and a configuration time budget $B$.

GGA starts by initializing a \emph{population} ($pop$) of configurations (named \emph{genomes}) as a subset of $\Theta$ in line~\ref{alg:gga:init-pop}.
This population is partitioned into a \emph{competitive} group (\emph{pop.comp}, which is directly evaluated on the target algorithm) and \emph{non-competitive} group (\emph{pop.noncomp}, which simply acts as a source of diversity).

The algorithm proceeds in a main loop that finishes when GGA reaches the configuration budget $B$ or a threshold on the performance (line~\ref{alg:gga:main-loop}).
At each iteration (which we call \emph{generation}), GGA selects a subset of the instances $\Pi_j$ to evaluate the genomes in line~\ref{alg:gga:inst-selection}\footnote{There are different policies that can be applied to select the instances at each generation, see~\cite{AnsoteguiST09}.}.
Then, in line~\ref{alg:gga:mini-tournament}, GGA evaluates the competitive genomes of the population over the selected instances $\Pi_j$ using a parallel racing scheme called \emph{mini-tournament}.
This procedure returns a set of $N$ winners, <$w_1, ..., w_{N}$>, which will be the only competitive genomes that will generate new offspring in this generation (line~\ref{alg:gga:offspring}).
Finally, GGA applies an ageing policy in line~\ref{alg:gga:aging} that is used to prevent population growth.
The only exception is the overall best competitive genome ($w_1$), which survives as long as it performs better than the other mini-tournament winners.
At the end of the main loop, GGA returns the best competitive genome $w_1$ of the last generation.

For more details on the GGA algorithm, we refer the reader to~\cite{AnsoteguiST09}.

\subsection{The SMAC Automatic Configurator}\label{sec:ac:smac}

Sequential Model-Based Algorithm Configuration (SMAC) is an automatic configuration algorithm based on Bayesian optimization~\cite{hutter_sequential_nodate,lindauer_smac3_2021}.
In Bayesian optimization, we use a few evaluations of the target algorithm to train a \emph{surrogate model} that predicts the performance of the algorithm for a given configuration.
This fast-to-evaluate surrogate model is used to search for promising new configurations that will be executed on the training instances.

\begin{algorithm}[ht]
    \begin{algorithmic}[1]
        \small
        \renewcommand{\algorithmicrequire}{\textbf{Input:}}
        
        \Require {Target Algorithm $A$, Parameter Space $\Theta$, Instances $\Pi$,
                  Performance Metric $\hat{c}$,
                  Configuration Budget $B$}
        \Function{SMAC}{$A, \Theta, \Pi, \hat{c}, B$}
        \State [$R$, $\theta_{inc}$] $\gets$ initialize($\Theta$, $\Pi$)\label{alg:smac:initialize}
        \While {$B$ not exhausted}\label{alg:smac:main-loop}
            \State [$M$, $t_{fit}$] $\gets$ fitModel(R)\label{alg:smac:fit-surrogate}
            \State [$\vec{\Theta}_{new}$, $t_{select}$] $\gets$ selectConfigurations($M$, $\theta_{inc}$, $\Theta$)\label{alg:smac:select-configs}
            \State [$R$, $\theta_{inc}$] $\gets$ intensify($A$, $\vec{\Theta}_{new}$, $\theta_{inc}$, $R$, $\Pi$, $\hat{c}$)\label{alg:smac:intensify}
        \EndWhile
        \Return $\theta_{inc}$
        \EndFunction
    \end{algorithmic}
    
    \caption{SMAC}
    \label{alg:smac}
\end{algorithm}

Algorithm~\ref{alg:smac} shows the pseudocode of SMAC.
This algorithm receives as input the target algorithm $A$, its parameter space $\Theta$, a set of training instances $\Pi$, a performance metric $\hat{c}$ to optimize and a configuration time budget $B$.
First, SMAC initializes a \emph{best candidate} configuration $\theta_{inc}$ and the history of conducted evaluations of different \emph{(configuration, instance)} pairs $R$ (which might be empty) in line~\ref{alg:smac:initialize}.

As in GGA (see Section~\ref{sec:ac:gga}), SMAC has a main loop defined in line~\ref{alg:smac:main-loop} that proceeds until the configuration budget $B$ is reached.
At each iteration, it fits a surrogate model $M$ using the information in $R$ in line~\ref{alg:smac:fit-surrogate}.
Then, it uses $M$ to select a new set of promising candidate configurations $\vec{\Theta}_{new}$ in line~\ref{alg:smac:select-configs}.
Finally, it evaluates $\vec{\Theta}_{new}$ and $\theta_{inc}$ on instances from $\Pi$ to determine the next best candidate $\theta_{inc}$, according to $\hat{c}$ in line~\ref{alg:smac:intensify}.
Similar to the GGA algorithm, SMAC returns the best candidate configuration $\theta_{inc}$ .

\subsection{Support for Tuning into the OptiLog framework}\label{sec:ac:optilog}

In this section, we present an excerpt of the code that uses the OptiLog framework to generate the configuration environment (from now on Tuning Scenario) of the solver Loandra for GGA and SMAC tuners.

\begin{lstlisting}[style=Python,label={lst:example_ac},caption={Sample code to wrap the solver Loandra into an OptiLog BlackBox.}]
# example_ac.py

from optilog.blackbox import *
from optilog.running import ParsingInfo
from optilog.tuning import *

class LoandraBB(SystemBlackBox):
    config = {
        "weight-strategy": Int(0, 2, default=2),
        "preprocess": Bool(default=True),
        (...)
    }
    (...)
\end{lstlisting}

Listing~\ref{lst:example_ac} defines a custom BlackBox class named LoandraBB that inherits from SystemBlackBox. This class represents the binary that we want to optimize. The \verb|config| dictionary defines the parameters of this binary that can be tuned by the optimization algorithm. We show the parameters ``weight-strategy'' and ``preprocess'', with their respective types and default values.

\begin{lstlisting}[style=Python,label={lst:ac_scenario_create},caption={Sample code to create a Tuning Scenario for the solver defined in Listing~\ref{lst:example_ac}.}]
# scenario_gga.py

from optilog.tuning.configurators import *
from example_ac import LoandraBB

if __name__ == "__main__":
    configurator = GGAConfigurator(
        LoandraBB(),
        input_data="/path/to/instances/*",
        ...
    )
    configurator.generate_scenario("./scenario")
\end{lstlisting}

Listing~\ref{lst:ac_scenario_create} is a definition of a Tuning Scenario for the solver Loandra. It imports the custom \verb|LoandraBB| class defined in Listing~\ref{lst:example_ac}, and sets up a tuner (in this case GGA) to optimize the parameters of \verb|LoandraBB|. The \verb|input_data| parameter specifies the path to the instances used during the optimization process. Lastly, the \verb|generate_scenario| method is called with the desired output path for the scenario that is being created.

Similar code to Listing~\ref{lst:ac_scenario_create} could be used to define a Tuning Scenario to be used with the SMAC AC tool, as OptiLog supports both GGA and SMAC.

\section{Configuring MaxSAT Solvers}\label{sec:configuring}

Although the AC tools (tuners) presented in the previous section have also parameters that impact the effectiveness of the configuration process, \emph{tuning the tuner} is out of reach in this paper and we focus on providing a good cost function to be used during the tuning process.

Ideally, we would use the $score(s, i)$ function from the MaxSAT Evaluation (Equation~\ref{eq:score_mse:score_inst}).
Notice though that in the MaxSAT Evaluation we are trying to maximize this scoring function, whereas tuners minimize a cost (see Section~\ref{sec:ac}).
Therefore, we have to convert the \emph{score} function to a \emph{cost} function.
Additionally, it is not guaranteed that the bounds found are equal or worse than the previously \emph{best-known upper bounds} (see Section~\ref{sec:mse}), and we cannot update the \emph{best-known upper bounds} sets during the tuning process (otherwise previous results computed in the same tuning process would not be comparable).

We define the $cost_{ac}(s, i)$ function, shown in Equation~\ref{eq:ac_ratio}, as follows.
First, we split the function in two cases: 1) the reported bound is worse (or equal) than the previous \emph{best-known upper bound}, and 2) the bound reported is better.

For 1), we compute $1 - score(s,i)$ to obtain a value in the range $[0, 1)$, where better bounds are closer to $0$.

For 2), notice that we are breaking the assumption $(best$-$known\ ub) \leq ub$, which may lead to unbounded values that tend to $\infty$.
To restrict the values to the range $(-1, 0)$, we use the inverse of the $score(s,i)$ function, and then subtract 1.

The $cost_{ac}(s,i)$ function returns values between $(-1, 1)$, where better bounds correspond to values closer to $-1$.

\begin{equation}\label{eq:ac_ratio}
cost_{ac}(s,i) =
\begin{cases}
    1 - score(s,i), &\mbox{if } \mbox{ub for }i \mbox{ found by }s \ge \mbox{best-known ub for } i\\
    \frac{1}{score(s, i)} - 1, &\mbox{otherwise} \\
\end{cases}
\end{equation}

\begin{figure}[ht]
    \centering
    \includegraphics[width=\textwidth]{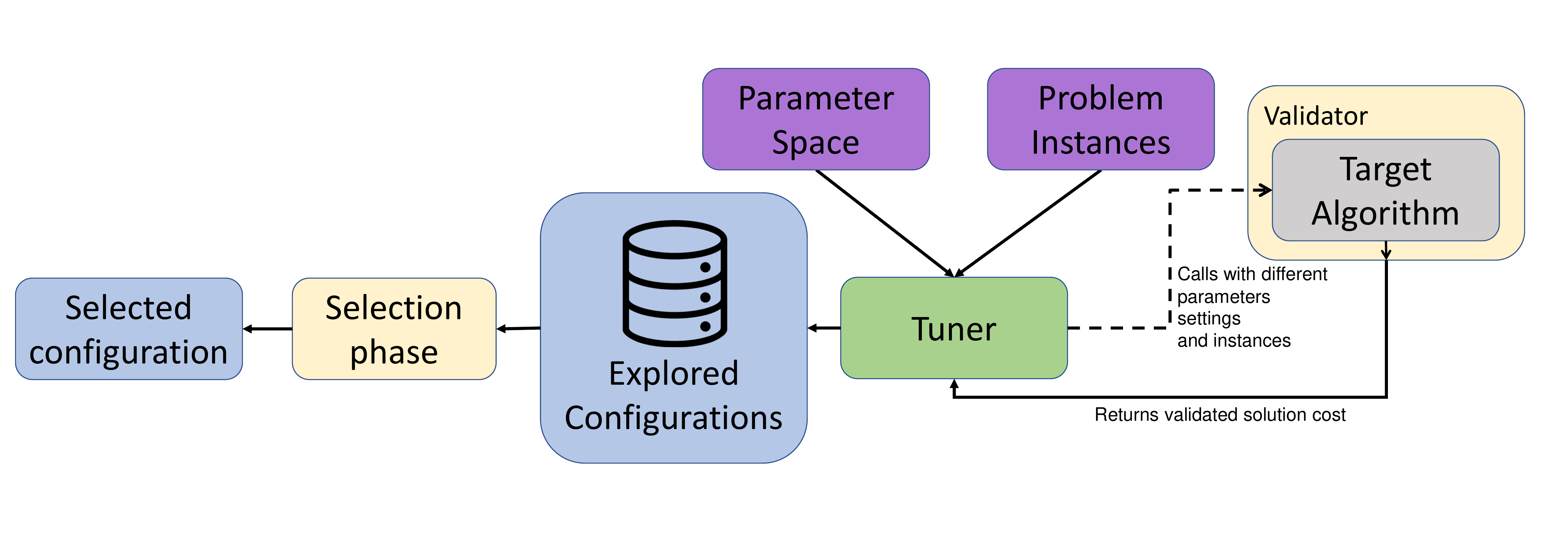}
    \caption{Visualization of the Automatic Configuration process extended with the validator and a selection phase over the explored configurations.}
    \label{fig2:aac-overview}
\end{figure}

As stated in Section~\ref{sec:mse:reproducing}, some executions might report a bound that does not match the reported solution. Thus, we integrate a validation step (see Figure~\ref{fig2:aac-overview}) 
that certifies the real cost of the solution returned by the solver and reports it to the tuner.

Regarding the tuning environment, all experiments are conducted on the same computation cluster. Each tuning process is given a wall-time tuning budget of 48 hours, a memory limit of 32G per worker, and is allowed to use up to 50 parallel workers unless otherwise specified. Each configuration instance is given a CPU time limit of 60 seconds for the solver, and then a validation step is executed. As training instances for the tuning process, we will use the instances from the MaxSAT Evaluation 2021 and we will test the best configuration returned by GGA and SMAC (see Section~\ref{sec:ac}) on the instances from the MaxSAT Evaluation 2022.

GGA allows for the selection of how many instances are used in each generation. Incrementally increasing the training set across generations till including the whole set of available instances is often recommended, as it facilitates discarding bad configurations with less effort, therefore more generations can be reached within the tuning budget. However, as discussed in Section~\ref{sec:discarded}, prioritizing the evaluation of more configurations on the whole training set within the same tuning budget may be preferable over having more generations. GGA also can preserve a set of \emph{elite} configurations that are run at every generation. We define as an \emph{elite} the default configuration of Loandra. Finally, we use the PyDGGA~\cite{Ansótegui2021} (version 1.7.0) distribution of GGA which has support for distributed execution.

Regarding SMAC, although it can be executed in parallel, it does not report an overall winner in contrast to GGA. Instead, it reports as many winners as computation cores were used since it basically runs several sequential SMACs in parallel. Thus, after SMAC completes, we have to take all the winners from each SMAC sequential execution, which may not have been evaluated on all training instances, and perform the missing evaluations. Then, the winner with the best performance on the training set is selected to be the overall winner. We use the SMAC3~\cite{lindauer_smac3_2021} (version 1.4.0) implementation of the algorithm.

As we have described earlier, the cost function used during the tuning process is not \emph{strictly} the minimization version of the score we maximize according to the MSE 2022 (see Section~\ref{sec:mse}). Therefore, one may argue that it would be better to return a winner for the training set with respect to the score function computed by the MSE. This is easy to do if the tuner provides the logs of each evaluation so, in the case of incomplete MaxSAT solvers, we can retrieve the best bound found by the solver on a given instance.

Therefore, we add a \emph{selection phase} (see Figure~\ref{fig2:aac-overview}) after the tuning phase that recomputes the scores (according to the MSE) of the configurations traversed by the tool during the tuning process.

In particular, for SMAC, we compute the MSE score of the 50 winners reported by the tuner on the training set and select the one with the highest score. Even though we have access to the logs of the evaluations of SMAC and could use those scores to select the winning configuration, we need to make sure that all the configurations are evaluated with all the instances.

For GGA, we order the configurations first by their ranking in a generation (according to the cost function in Equation~\ref{sec:configuring}), and within the same rank, we order by the most recent generation. Then, we select the first 50 distinct configurations. We look into their logs, recompute their MSE score according to Section~\ref{sec:mse}, and report the winner\footnote{In our experiments, the winner reported by GGA was the same configuration as the best one found in the \emph{selection phase}.}.

\begin{table}[ht]
\centering
\caption{Comparison using GGA and SMAC to tune the Loandra solver (using $VBS_b + MSE_b + LRUNS_b$ bounds).}
\label{tab:gga_vs_smac}
\begin{tabular}{l|rrrrr}
\toprule
 & Mean & Median & Min & Max & Std \\
\midrule
NuWLS-c & 0.7524 & 0.7522 & 0.7484 & 0.7560 & 0.0017 \\
Loa (GGA, all-i) & 0.7393 & 0.7391 & 0.7313 & 0.7475 & 0.0037 \\
Loa (GGA, incremental) & 0.7353 & 0.7354 & 0.7275 & 0.7433 & 0.0038 \\
DT-Hywalk & 0.7351 & 0.7355 & 0.7288 & 0.7415 & 0.0030 \\
Loa (SMAC) & 0.7237 & 0.7234 & 0.7149 & 0.7355 & 0.0048 \\
TT-Open-WBO-inc (g) & 0.7164 & 0.7165 & 0.7128 & 0.7194 & 0.0015 \\
TT-Open-WBO-inc (i) & 0.7141 & 0.7142 & 0.7093 & 0.7188 & 0.0020 \\
TT-Open-WBO-inc (is) & 0.7118 & 0.7117 & 0.7098 & 0.7145 & 0.0008 \\
Loandra & 0.6953 & 0.6957 & 0.6872 & 0.7036 & 0.0037 \\
\bottomrule
\end{tabular}
\end{table}

Table~\ref{tab:gga_vs_smac} shows the result of the best configurations provided by GGA using all the instances from the first generation (``\emph{Loa (GGA, all-i)}''), or adding them incrementally at each step (``\emph{Loa (GGA, incremental)}'') and the best configuration provided by SMAC (see ``\emph{Loa (SMAC)}'') after the additional selections process described in the paragraph above. For the incremental approach of GGA, we use 20\% of instances at the first generation and instruct GGA to use all the instances on generation 25. Those values were selected based on preliminary experiments taking into account the number of generations that GGA can do in the given time.
It is clear by the results that the usage of all the instances from the beginning benefits GGA, allowing it to lift Loandra from the sixth position to the second one. In the next section, we will focus on the variant of GGA ``\emph{Loa (GGA, all-i)}''.

\section{Exploiting Configurations \emph{Discarded} by the Tuner}\label{sec:discarded}

As it has been shown in the literature~\cite{hamadi_manysat_2009, olivier_roussel_description_nodate}, from the most pragmatic point of view, we can obtain an efficient parallel approach by just running the same non-deterministic solver with different seeds in parallel, or we can also run in parallel different configurations of the same solver. 

In case resources are limited, we can also schedule the execution of different configurations of the same solver.
In this section, we concrete and study these different approaches.
We use OptiLog~\cite{alos_optilog_2022} to generate all the portfolios, as we explain in Section~\ref{sec:optilog:porfolio-gen}.

\subsection{Parallel Portfolios of seeds and configurations}\label{sec:portfolios:parallel}

As we have already explained, tuners report the best configuration they have found.
However, many other \emph{potentially} good configurations are also explored and discarded during the automatic configuration process with respect to their performance on the particular training set.
These configurations may exhibit good performance in different kinds of instances.
As observed in \cite{ansotegui_boosting_2022} on SAT benchmarks, superior performance can be achieved by combining these complementary configurations.

The first approach we explored is the parallel execution of $N$ different random seeds over a given MaxSAT solver.
This approach can be applied to both the \emph{default} MaxSAT solver and the best configuration obtained in the tuner.

Another approach is to extract $N$ configurations of a MaxSAT solver from the ones traversed by the tuner and execute them in parallel. 
There are many strategies that we could follow to extract these configurations from the tuner. In particular, we use the set of configurations considered during the \emph{selection phase} (after the \emph{tuning phase}) process as explained in Section~\ref{sec:configuring}.
Notice that we do not analyze any structure of the instances and we only incorporate configurations of the same solver.

Table~\ref{tab:parallel:all} shows the results of the parallel portfolios that we explained.
We tested parallel portfolios with 25, 30, 35, 40, 45, and 50 parallel executions.
Each row shows the results of a parallel portfolio (rows marked with \emph{(Seeds)} refer to a parallel portfolio of seeds, whereas the row marked with \emph{(Configs)} refer to a parallel portfolio of configurations).
We show the score as computed in the MaxSAT evaluation using the $VBS_b + MSE_b + LRUNS_b$ upper bounds and the rank of each portfolio with respect to the others.

As we can observe, the portfolio over different seeds for the default Loandra  (``\emph{(Seeds) Loandra}'' in Table~\ref{tab:parallel:all}) is not competitive while the portfolio of different seeds for the best configuration of Loandra computed by GGA (column ``\emph{(Seeds) Loa (GGA, all-i)}'') already outperforms NuWLS-c. Additionally, a portfolio of the best configurations provided by the selection phase (column ``\emph{(Configs) Loa (GGA, all-i)}'') systematically outperforms the rest of the approaches. These observations hold almost for any number of parallel executions.

\begin{table}[ht]
\centering
\caption{Score and rank (\#) for each parallel portfolio, given N parallel processes (using $VBS_b + MSE_b + LRUNS_b$ bounds).}
\label{tab:parallel:all}

\resizebox{\columnwidth}{!}{\begin{tabular}{l|llllllllllll}
\toprule
$N$ & \multicolumn{2}{r}{25} & \multicolumn{2}{r}{30} & \multicolumn{2}{r}{35} & \multicolumn{2}{r}{40} & \multicolumn{2}{r}{45} & \multicolumn{2}{r}{50} \\
                                         & \# & score & \# & score & \# & score & \# & score & \# & score & \# & score \\
\midrule
(Configs) Loa (GGA, all-i)               & 1 & 0.813592 & 1 & 0.818663 & 1 & 0.823986 & 1 & 0.825208 & 1 & 0.826448 & 1 & 0.827105 \\
(Seeds) Loa (GGA, all-i)                 & 2 & 0.806889 & 2 & 0.808078 & 2 & 0.809922 & 2 & 0.812589 & 2 & 0.813007 & 2 & 0.815263 \\
(Seeds) NuWLS-c                          & 3 & 0.768409 & 3 & 0.769302 & 3 & 0.769785 & 3 & 0.770293 & 3 & 0.771219 & 3 & 0.771263 \\
(Seeds) DT-Hywalk                        & 4 & 0.759271 & 4 & 0.764688 & 4 & 0.765736 & 4 & 0.765764 & 4 & 0.765862 & 4 & 0.766397 \\
(Seeds) TT-Open-WBO-inc (g)              & 5 & 0.728901 & 5 & 0.728903 & 5 & 0.729766 & 5 & 0.729902 & 5 & 0.730165 & 5 & 0.730235 \\
(Seeds) TT-Open-WBO-inc (i)              & 6 & 0.725971 & 6 & 0.726171 & 6 & 0.726253 & 6 & 0.726315 & 6 & 0.726510 & 6 & 0.727468 \\
(Seeds) Loandra                          & 8 & 0.717788 & 8 & 0.722663 & 7 & 0.723954 & 7 & 0.723996 & 7 & 0.724521 & 7 & 0.724648 \\
(Seeds) TT-Open-WBO-inc (is)             & 7 & 0.722672 & 7 & 0.722704 & 8 & 0.722918 & 8 & 0.723174 & 8 & 0.723324 & 8 & 0.723489 \\
\bottomrule
\end{tabular}}
\end{table}

\subsection{Sequential Portfolios of configurations}\label{sec:portfolios:sequential}

In some settings, we will not have enough resources to run a parallel portfolio as described in Section~\ref{sec:portfolios:parallel}. Potentially, we can have just one computation core available. In this case, we can schedule the sequential execution of different configurations of Loandra within the given timeout.

Let us describe how we construct this \emph{sequential} portfolio. We assume we have a sequence of solvers (or configurations of a solver) ($S$) that iteratively report better solutions, a time budget ($TO$), and a maximum time budget a solver can exhaust between two consecutive reported solutions ($MTBS$). The solvers are executed according to their order in the sequence until the time consumed globally by all the solvers exceeds $TO$.

Each solver is run as follows: first, we wait for the first solution reported by the solver. Once this first solution is reported, we start a timer of $MTBS$ seconds. If the solver reports a new solution before this timer expires, we reset the timer and wait for a new solution. This is repeated until the solver is unable to report a new solution before the timer is consumed. At that point in time, the solver is stopped and the next one in the ordered list of solvers is executed. Note that, at any point in this process, a solver can also be stopped if the global time budget of $TO$ seconds gets exhausted. A special case is the last solver of the sequence, which is allowed to run until the time budget expires (i.e. it is not stopped even if it took more than $MTBS$ seconds to find a new solution). Obviously, we keep track of the best overall solution seen so far.

To identify which sequence of solvers $S$ and $MTBS$ value the portfolio should use, we carry out a simulation of sequential portfolios with the configurations provided by the \emph{selection phase} (see Section~\ref{sec:configuring}) and their respective logs on the training instances computed during the \emph{tuning phase}. In particular, we explore all sequences of up to size 3 and $MTBS$ values of $\{2, 3, 5, 10, 15, 20, 25, 30, 35, 40, 45, 50\}$ seconds. Once we identify the best \emph{virtual} sequential portfolio for the training instances, we simulate again the execution of this \emph{virtual} sequential portfolio on the test set. In Table~\ref{tab:all:seq_theoric} we present the results of this simulation.

To implement this \emph{virtual} sequential portfolio we would need to take into account an additional thread that keeps track of the evolution of the solvers in the sequence, which may decrease the overall performance. Therefore, we see this \emph{virtual} sequential portfolio as a restarting policy that MaxSAT developers could integrate into their solvers, with the added benefit that they may be able to reuse information computed by each solver in the sequence.

\begin{table}[t]
\centering
\caption{Score of the \emph{virtual} sequential portfolio compared with the single-execution approach (using $VBS_b + MSE_b + LRUNS_b$ bounds).}
\label{tab:all:seq_theoric}
\begin{tabular}{l|r}
\toprule
 & Score \\
\midrule
Virtual sequential portfolio (N=2) - Loa (gga, all-i) & 0.7642 \\
Virtual sequential portfolio (N=3) - Loa (gga, all-i) & 0.7642 \\
NuWLS-c (max score on 50 seeds) & 0.7560 \\
NuWLS-c & 0.7554 \\
Loa (gga, all-i) & 0.7513 \\
DT-Hywalk & 0.7432 \\
TT-Open-WBO-inc (g) & 0.7214 \\
TT-Open-WBO-inc (i) & 0.7180 \\
TT-Open-WBO-inc (is) & 0.7180 \\
Loandra & 0.6965 \\
\bottomrule
\end{tabular}
\end{table}

Table~\ref{tab:all:seq_theoric} shows the results of the \emph{virtual} sequential portfolios (rows prefixed with ``Virtual portfolio''), compared to the results that obtained the solvers from the competition with the default parameters, and with the best approach obtained using a tuner (``\emph{Loa (GGA, all-i)}''). As in the MSE we run each solver with the same seed, except for NuWLS-c for which we also report on the best score value from 50 seeds. The $N$ value shown in the \emph{virtual} sequential portfolios rows indicates the length of the solvers' sequence. The portfolios are built on top of the configurations obtained after the selection phase with (``\emph{Loa (GGA, all-i)}'').

We notice that \emph{virtual} sequential portfolios do perform better than NuWLS-c, and a selection of two configurations suffices to that end. Interestingly, if we build the \emph{virtual} sequential portfolio on the test instances from the MSE 2022, then we get a better portfolio using three configurations that achieves a score of $0.7689$, however, we cannot predict this portfolio based on the analysis we perform on the training instances from the MSE 2021.

\subsection{OptiLog Portfolio Generator}\label{sec:optilog:porfolio-gen}

To facilitate the creation of the parallel and \emph{virtual} sequential portfolios, we added support to compute them using the OptiLog framework.

\begin{lstlisting}[style=Python,caption={Computing a parallel portfolio with OptiLog.},label={lst:parallel_portfolio}]
from optilog.portfolio import get_parallel_portfolio

get_parallel_portfolio(
    gga_scenario='./tuning-scenario',
    n_solvers=10,
    save_to='./parallel-portfolio'
)
\end{lstlisting}

Listing~\ref{lst:parallel_portfolio} shows how we can generate and save a parallel portfolio with OptiLog. This portfolio is built by selecting $N$ configurations as explained in Section~\ref{sec:configuring}, thus requiring a Tuning Scenario (generated with OptiLog as seen in Section~\ref{sec:optilog:porfolio-gen}). The function \verb|get_parallel_portfolio| receives as parameters the Tuning Scenario that contains the results of the tuning process (\verb|gga_scenario|), the number of solvers that will compose the parallel portfolio (\verb|n_solvers|), and the directory where the scripts to launch each individual solver that composes the portfolio will be saved (\verb|save_to|).

\begin{lstlisting}[style=Python,caption={Computing a sequential portfolio with OptiLog.},label={lst:seq_portfolio}]
from optilog.portfolio import get_sequential_portfolio

get_sequential_portfolio(
    path_scenario="./running-scenario",
    n_solvers=2,
    solution_regex=r"^o\s(\d+)",
    save_to="./sequential-portfolio",
    score_fn=maxsat_score_fn,
    max_time_between_solutions=[5, 15, 25, 35]
)
\end{lstlisting}

Listing~\ref{lst:seq_portfolio} shows how we are generating a sequential portfolio with the results of a Running Scenario. Note that to generate the \emph{virtual} sequential portfolio we require the full trace of the solvers (in particular for the incomplete MaxSAT case, we need the evolution of the best bound over time), so we cannot build it from a Tuning Scenario directly. The parameters \verb|gga_scenario|, \verb|save_to|, and \verb|n_solvers| mean the same as in the function to compute a parallel portfolio. Additionally, we have to specify the following parameters: \verb|score_fn| is used to transform the lines matched by \verb|solution_regex| to a score that the portfolio will try to maximize (in this example the score function is $score(s)$ defined in Section~\ref{sec:mse}), and \verb|max_time_between_solutions| contains the possible values that the portfolio can choose from when selecting the parameter $MTBS$.

\section{Conclusions}

Given a target solver, we have presented an approach to easily generate a potentially much better solving approach. To this end, we exploit a set of alternative configurations of the same target solver coming from the residues of a tuning process. It is important to notice that we do not exploit any structure feature of the input problem or instance since in some domains these features are not easy to compute. In particular, we have shown how from a MaxSAT solver with a low ranking in one of the tracks of the MSE 2022 we can obtain a more competitive approach.

Our sequential portfolio generation approach can be seen as a first attempt to come up with effective restarting policies for MaxSAT solvers, something that has not been studied in depth in the literature.

Finally, the approach described has been integrated into the OptiLog framework avoiding the tedious process of setting up tuning environments and generating portfolios. Moreover, the API is general enough to be applied not only to MaxSAT solvers but to other solving approaches.

\bibliographystyle{unsrtnat}
\bibliography{bibliography}

\begin{thebibliography}{18}
\providecommand{\natexlab}[1]{#1}
\providecommand{\url}[1]{\texttt{#1}}
\expandafter\ifx\csname urlstyle\endcsname\relax
  \providecommand{\doi}[1]{doi: #1}\else
  \providecommand{\doi}{doi: \begingroup \urlstyle{rm}\Url}\fi

\bibitem[Bacchus et~al.(2022)Bacchus, Berg, Järvisalo, Martins, and
  Niskanen]{bacchus_maxsat_2022}
Fahiem Bacchus, Jeremias Berg, Matti Järvisalo, Ruben Martins, and Andreas
  Niskanen.
\newblock {MaxSAT} {Evaluation} 2022 : {Solver} and {Benchmark} {Descriptions}.
\newblock \emph{Department of Computer Science Series of Publications B},
  B-2022-2, 2022.
\newblock URL \url{https://helda.helsinki.fi/handle/10138/347396}.
\newblock Accepted: 2022-08-25T10:09:01Z Publisher: Department of Computer
  Science, University of Helsinki.

\bibitem[Ansótegui et~al.(2009)Ansótegui, Sellmann, and
  Tierney]{AnsoteguiST09}
Carlos Ansótegui, Meinolf Sellmann, and Kevin Tierney.
\newblock A {Gender}-based {Genetic} {Algorithm} for the {Automatic}
  {Configuration} of {Algorithms}.
\newblock In \emph{Proceedings of the 15th {International} {Conference} on
  {Principles} and {Practice} of {Constraint} {Programming}}, {CP}'09, pages
  142--157. Springer-Verlag, 2009.
\newblock ISBN 3-642-04243-0 978-3-642-04243-0.
\newblock URL \url{http://dl.acm.org/citation.cfm?id=1788994.1789011}.
\newblock tex.acmid: 1789011 tex.numpages: 16 tex.year: 2009 event-place:
  Lisbon, Portugal.

\bibitem[Lindauer et~al.(2021)Lindauer, Eggensperger, Feurer, Biedenkapp, Deng,
  Benjamins, Ruhkopf, Sass, and Hutter]{lindauer_smac3_2021}
Marius Lindauer, Katharina Eggensperger, Matthias Feurer, André Biedenkapp,
  Difan Deng, Carolin Benjamins, Tim Ruhkopf, René Sass, and Frank Hutter.
\newblock {SMAC3}: {A} {Versatile} {Bayesian} {Optimization} {Package} for
  {Hyperparameter} {Optimization}.
\newblock In \emph{{ArXiv}: 2109.09831}, 2021.
\newblock URL \url{https://arxiv.org/abs/2109.09831}.

\bibitem[Kadioglu et~al.(2010)Kadioglu, Malitsky, Sellmann, and
  Tierney]{kadioglu_isac_2010}
Serdar Kadioglu, Yuri Malitsky, Meinolf Sellmann, and Kevin Tierney.
\newblock {ISAC} – {Instance}-{Specific} {Algorithm} {Configuration}.
\newblock In \emph{{ECAI} 2010}, pages 751--756. IOS Press, 2010.
\newblock \doi{10.3233/978-1-60750-606-5-751}.
\newblock URL
  \url{https://ebooks.iospress.nl/doi/10.3233/978-1-60750-606-5-751}.

\bibitem[Alòs et~al.(2022)Alòs, Ansótegui, Salvia, and
  Torres]{alos_optilog_2022}
Josep Alòs, Carlos Ansótegui, Josep~M. Salvia, and Eduard Torres.
\newblock {OptiLog} {V2}: {Model}, {Solve}, {Tune} and {Run}.
\newblock In Kuldeep~S. Meel and Ofer Strichman, editors, \emph{25th
  {International} {Conference} on {Theory} and {Applications} of
  {Satisfiability} {Testing} ({SAT} 2022)}, volume 236 of \emph{Leibniz
  {International} {Proceedings} in {Informatics} ({LIPIcs})}, pages
  25:1--25:16, Dagstuhl, Germany, 2022. Schloss Dagstuhl – Leibniz-Zentrum
  für Informatik.
\newblock ISBN 978-3-95977-242-6.
\newblock \doi{10.4230/LIPIcs.SAT.2022.25}.
\newblock URL \url{https://drops.dagstuhl.de/opus/volltexte/2022/16699}.
\newblock ISSN: 1868-8969.

\bibitem[Berg et~al.(2019)Berg, Demirovic, and Stuckey]{berg_loandra_2019}
Jeremias Berg, Emir Demirovic, and Peter Stuckey.
\newblock Core-boosted linear search for incomplete maxsat.
\newblock \emph{Integration of Constraint Programming, Artificial Intelligence,
  and Operations Research: 16th International Conference, CPAIOR 2019}, 2019.

\bibitem[Zheng et~al.(2022)Zheng, He, Chen, Zhou, and Li]{zheng2022decision}
Jiongzhi Zheng, Kun He, Zhuo Chen, Jianrong Zhou, and Chu-Min Li.
\newblock Decision tree based hybrid walking strategies.
\newblock \emph{MaxSAT Evaluation 2022}, page~24, 2022.

\bibitem[L{\"u}bke and Schupp(2022)]{lubke2022nosat}
Ole L{\"u}bke and Sibylle Schupp.
\newblock nosat-maxsat.
\newblock In \emph{MaxSAT Evaluation 2022}, pages 29--30. Department of
  Computer Science, University of Helsinki, 2022.

\bibitem[Chu et~al.(2022)Chu, Cai, Lei, and He]{chu2022nuwls}
Yi~Chu, Shaowei Cai, Zhendong Lei, and Xiang He.
\newblock Nuwls-c: Solver description.
\newblock \emph{MaxSAT Evaluation 2022}, page~28, 2022.

\bibitem[Elffers and Nordström(2018)]{elffers_divide_2018}
Jan Elffers and Jakob Nordström.
\newblock Divide and conquer: Towards faster pseudo-boolean solving.
\newblock \emph{Proceedings of the Twenty-Seventh International Joint
  Conference on Artificial Intelligence}, pages 1291--1299, 2018.
\newblock URL \url{https://www.ijcai.org/Proceedings/2018/180}.

\bibitem[{Jo Devriendt}(2023)]{jo_devriendt_exact_2023}
{Jo Devriendt}.
\newblock Exact {Solver} {Repository}, April 2023.
\newblock URL \url{https://gitlab.com/JoD/exact}.

\bibitem[Joshi et~al.(2019)Joshi, Kumar, Rao, and
  Martins]{DBLP:journals/jsat/JoshiKRM19}
Saurabh Joshi, Prateek Kumar, Sukrut Rao, and Ruben Martins.
\newblock Open-wbo-inc: Approximation strategies for incomplete weighted
  maxsat.
\newblock \emph{J. Satisf. Boolean Model. Comput.}, 11\penalty0 (1):\penalty0
  73--97, 2019.
\newblock \doi{10.3233/SAT190118}.
\newblock URL \url{https://doi.org/10.3233/SAT190118}.

\bibitem[Nadel(2020)]{nadel_polarity_2020}
Alexander Nadel.
\newblock Polarity and {Variable} {Selection} {Heuristics} for {SAT}-{Based}
  {Anytime} {MaxSAT}: {System} {Description}.
\newblock \emph{Journal on Satisfiability, Boolean Modeling and Computation},
  12\penalty0 (1):\penalty0 17--22, September 2020.
\newblock ISSN 15740617.
\newblock \doi{10.3233/SAT-200126}.
\newblock URL
  \url{https://www.medra.org/servlet/aliasResolver?alias=iospress&doi=10.3233/SAT-200126}.

\bibitem[Hutter et~al.(2011)Hutter, Hoos, and
  Leyton-Brown]{hutter_sequential_nodate}
Frank Hutter, Holger~H Hoos, and Kevin Leyton-Brown.
\newblock Sequential {Model}-{Based} {Optimization} for {General} {Algorithm}
  {Conﬁguration} (extended version).
\newblock \emph{International Conference on Learning and Intelligent
  Optimization}, 2011.

\bibitem[Ans{\'o}tegui et~al.(2021)Ans{\'o}tegui, Pon, and
  Sellmann]{Ansótegui2021}
Carlos Ans{\'o}tegui, Josep Pon, and Meinolf Sellmann.
\newblock Boosting evolutionary algorithm configuration.
\newblock \emph{Annals of Mathematics and Artificial Intelligence}, 2021.
\newblock ISSN 1573-7470.
\newblock \doi{10.1007/s10472-020-09726-y}.
\newblock URL \url{https://doi.org/10.1007/s10472-020-09726-y}.

\bibitem[Hamadi et~al.(2009)Hamadi, Jabbour, and Sais]{hamadi_manysat_2009}
Youssef Hamadi, Said Jabbour, and Lakhdar Sais.
\newblock {ManySAT}: a parallel {SAT} solver.
\newblock \emph{JSAT}, 6:\penalty0 245--262, June 2009.
\newblock \doi{10.3233/SAT190070}.

\bibitem[{Olivier Roussel}(2012)]{olivier_roussel_description_nodate}
{Olivier Roussel}.
\newblock Description of ppfolio 2012.
\newblock In \emph{Proceedings of {SAT} {Challenge} 2012: {Solver} and
  {Benchmark} {Descriptions}}, 2012.

\bibitem[Ansótegui et~al.(2022)Ansótegui, Pon, and
  Sellmann]{ansotegui_boosting_2022}
Carlos Ansótegui, Josep Pon, and Meinolf Sellmann.
\newblock Boosting evolutionary algorithm configuration.
\newblock \emph{Annals of Mathematics and Artificial Intelligence}, 90\penalty0
  (7-9):\penalty0 715--734, 2022.
\newblock Publisher: Springer.

\end{thebibliography}
\end{document}